\documentclass[runningheads]{llncs}

\usepackage{eccv}

\usepackage{eccvabbrv}

\usepackage{graphicx}
\usepackage{booktabs}
\usepackage{multirow}
\usepackage{amssymb}
\usepackage{bbding}
\usepackage{makecell}
\usepackage[dvipsnames]{xcolor}
\usepackage[accsupp]{axessibility}  

\usepackage{hyperref}

\usepackage{orcidlink}

\makeatletter
\renewcommand*{\@fnsymbol}[1]{\ifcase#1\or\dag\or\ddag\or*\or**\or***\or****\or*****\or******\or*******\or********\or*********\else\@ctrerr\fi}
\makeatother

\begin{document}

\title{XPSR: Cross-modal Priors for Diffusion-based Image Super-Resolution} 

\titlerunning{XPSR}

\author{Yunpeng Qu\inst{1,2}\thanks{Equal contribution. }\orcidlink{0009-0006-9700-6290}, Kun Yuan\inst{2}$^{\dag}$\thanks{Project leader.}\orcidlink{0000−0002−3681−2196}, Kai Zhao\inst{2}\orcidlink{0009-0003-5237-7512
}, \\
Qizhi Xie\inst{1,2}\orcidlink{0000-0001-6171-9789}, Jinhua Hao\inst{2}, Ming Sun\inst{2} \and Chao Zhou\inst{2}}

\authorrunning{Y Qu et al.}

\institute{Tsinghua University, China, Beijing \and Kuaishou Technology, China, Beijing\\
\email{\{qyp21,xqz20\}@mails.tsinghua.edu.cn}\\
\email{\{yuankun03,zhaokai05,haojinhua,sunming03,zhouchao\}@kuaishou.com}}

\maketitle

\begin{abstract}
Diffusion-based methods, endowed with a formidable generative prior, have received increasing attention in Image Super-Resolution (ISR) recently. 
However, as low-resolution (LR) images often undergo severe degradation, it is challenging for ISR models to perceive the semantic and degradation information, resulting in restoration images with incorrect content or unrealistic artifacts.
To address these issues, we propose a \textit{Cross-modal Priors for Super-Resolution (XPSR)} framework.
Within XPSR, to acquire precise and comprehensive semantic conditions for the diffusion model, cutting-edge Multimodal Large Language Models (MLLMs) are utilized.
To facilitate better fusion of cross-modal priors, a \textit{Semantic-Fusion Attention} is raised.
To distill semantic-preserved information instead of undesired degradations, a \textit{Degradation-Free Constraint} is attached between LR and its high-resolution (HR) counterpart.
Quantitative and qualitative results show that XPSR is capable of generating high-fidelity and high-realism images across synthetic and real-world datasets.
Codes are released at \url{https://github.com/qyp2000/XPSR}.
  \keywords{Image super-resolution \and Image restoration \and Diffusion models \and Multimodal large language models}
\end{abstract}

\section{Introduction}
\label{sec:intro}
The objective of Image Super-Resolution (ISR) entails generating a perceptually authentic how-resolution (HR) image from its low-resolution (LR) counterpart, which is characterized by unknown and intricate degradation processes. Despite the considerable achievements of GAN-based methods \cite{DBLP:conf/iccv/0008LGT21, DBLP:conf/iccvw/WangXDS21, DBLP:conf/iccvw/LiangCSZGT21}, they still struggle to reproduce vivid and realistic textures. This is mainly attributable to the domain gap between the synthetic training data and the real-world test data, and the excessive fidelity-oriented optimization objective \cite{DBLP:journals/corr/abs-2311-16518}. 

Recently, denoising diffusion probabilistic models (DDPMs) are emerging as successors to GANs across a range of generation tasks \cite{ho2020denoising, rombach2022high, zhang2023adding, DBLP:journals/corr/abs-2302-08453}, due to their strong ability in modeling complicated distributions. Some pioneering works \cite{DBLP:conf/nips/KawarEES22, DBLP:conf/cvpr/FeiLPZYLZ023} have adopted DDPMs to tackle the ISR problems. Notably, ISR introduces further challenges, on account of high restoration fidelity. This requirement starkly contrasts with the inherent stochasticity of DDPMs. 

Given that the large-scale pretrained text-to-image (T2I) models (\eg StableDiffusion (SD) \cite{rombach2022high}), trained on datasets surpassing 5 billion image-text pairs and embodying robust and plentiful natural image priors, other methods \cite{DBLP:journals/corr/abs-2305-07015, DBLP:journals/corr/abs-2308-15070, yang2023pasd, DBLP:journals/corr/abs-2311-16518} utilize ControlNet \cite{zhang2023adding} to harness pre-trained SD priors for ISR. Encouragingly, these methods have exhibited the incredible capability to generate realistic image details. 
StableSR \cite{DBLP:journals/corr/abs-2305-07015} and DiffBIR \cite{DBLP:journals/corr/abs-2308-15070} straightforwardly set the prompt condition to empty, relying on extracting semantic information from LR images.
However, LR images undergo complex degradations, making it challenging for ISR models to extract semantic priors, which are provided in the form of textual prompts within pre-trained T2I models.
PASD \cite{yang2023pasd} and SeeSR \cite{DBLP:journals/corr/abs-2311-16518} leverage off-the-shelf tagging models for the extraction of object labels as high-level prompts.
However, these tagging prompts lack more complex information such as spatial location and scene understanding, which are crucial for generating comprehensive images in T2I models.
These prompts also fail to capture the inherent distortions within images, yet such low-level priors facilitate the modeling of the degradation process, thereby enabling clearer restorations in ISR \cite{zhang2018learning,bell2019blind}.

In this paper, we explore the significance of different semantic priors from multi-modal large language models (MLLMs) for ISR. 
Based on the analysis, we present a \textit{Cross-modal Priors for Super-Resolution (XPSR)} framework, utilizing cross-modal priors to guide diffusion models in generating more high-fidelity and realistic images.
To furnish more precise and perceptually aligned semantic priors, we introduce MLLMs to offer multi-level prompts of LR images.
To effectively fuse cross-model semantic priors with generative priors, we propose a \textit{Semantic-Fusion Attention (SFA)} for adaptive selection of semantic features.
To enhance the extraction of semantic-preserved features, we further devise a \textit{Degradation-Free Constraint (DFC)}.
Our \textbf{contributions} are as follows:
\begin{itemize}
    \item[1.] We explore the significance of semantic priors in diffusion-based ISR, wherein high-level priors offer a wealth of semantic information while low-level priors assist in modeling degradation mechanisms. Furthermore, cutting-edge MLLMs are utilized to obtain the desired appropriate priors.
    \item[2.] We propose the XPSR, where the SFA is adopted to fuse multi-level semantic priors with the diffusion model in a parallel cross-attention manner. 
    \item[3.] To obtain semantic-preserved features instead of degradations, the DFC is attached between the LR and HR images in the pixel and latent space.
    \item[4.] Through quantitative and qualitative analysis, XPSR demonstrates a strong capability in generating high-fidelity and high-realism images, achieving \textbf{the best performance} on multiple image quality metrics across several datasets. Extensive ablation studies prove the validity of each component.

\end{itemize}

\section{Related Work}
\label{sec:related}

\subsection{Degradation Prior-based Image Super-resolution}

Initial investigations \cite{DBLP:journals/corr/abs-2311-14282, dai2019second, dong2015image, DBLP:conf/eccv/DongLT16, DBLP:conf/cvpr/LimSKNL17,DBLP:conf/eccv/ZhangLLWZF18} into ISR mainly focused on the restoration of LR images through pre-determined degradation types, such as bicubic downsampling, blurring, noise, and other factors. Yet, their efficacy markedly diminishes in practical settings due to their limited adaptability. \cite{DBLP:conf/cvpr/GuLZD19, DBLP:conf/iccv/MichaeliI13, DBLP:conf/icml/ZhangGCDZY23}.
To elevate performance further, recent advancements (\ie, GAN-based methods \cite{DBLP:conf/iccvw/LiangCSZGT21, chen2022real,DBLP:conf/iccv/0008LGT21,DBLP:conf/iccvw/WangXDS21,DBLP:conf/cvpr/LiangZZ22}) have delved into more complex degradation models to mimic the real-world distortions closely. Among them, BSRGAN \cite{DBLP:conf/iccv/0008LGT21} proposed using random shuffling combinations of basic degradation operations. Real-ESRGAN \cite{DBLP:conf/iccvw/WangXDS21} employed a high-order process to simulate degradation in real-life procedures, \eg, cameras, image editing, transmission, \etc.
Beyond the dependence on explicit priors, other studies \cite{DBLP:conf/cvpr/MenonDHRR20,DBLP:conf/cvpr/GuSZ20,DBLP:conf/cvpr/ChanWXGL21, DBLP:conf/aaai/LiZQLL22, luo2024and} also discovered adopting implicit priors through adversarial training. However, these methods fall short, chiefly due to insufficient prior knowledge, resulting in suboptimal output quality.

\subsection{Diffusion Prior-based Image Super-resolution}

Analogous to GAN-based methodologies, diffusion models \cite{DBLP:conf/nips/HoJA20,DBLP:conf/cvpr/RombachBLEO22,DBLP:journals/corr/abs-2302-08453, saharia2022photorealistic, zhang2023adding, ramesh2022hierarchical, sahak2023denoising} have earned escalating recognition for their superior ability to embody generative priors. Of late, these models have been skillfully utilized as generative priors within the realm of ISR as well \cite{saharia2022image, DBLP:journals/pami/SahariaHCSFN23, kawar2022denoising, wang2023exploiting}. Among them, StableSR \cite{DBLP:journals/corr/abs-2305-07015} trained a time-aware encoder to fine-tune the StableDiffusion (SD) model \cite{rombach2022high} and employed feature warping to balance between fidelity and perceptual quality. DiffBIR \cite{DBLP:journals/corr/abs-2308-15070} utilized a two-stage approach, initially reconstructing the image as a preliminary estimate, and subsequently leveraging the SD to refine and augment visual details. To effectively utilize the potential of pretrained text-to-image diffusion models, PASD \cite{yang2023pasd} and SeeSR \cite{DBLP:journals/corr/abs-2311-16518} introduced off-the-shelf high-level semantic information (\ie, object tags) as extra conditions. Through these endeavors, ISR can reproduce more realistic image details. Nevertheless, when confronted with LR images that exhibit complicated distortions, and intertwined spatial positions among objects, the generated result will be substandard.

\subsection{Cross-modal Semantic Prior-based Image Super-resolution}
Textual prompts are essential in guiding targeted image generation in T2I diffusion models \cite{DBLP:conf/cvpr/BrooksHE23, DBLP:conf/cvpr/KawarZLTCDMI23, DBLP:conf/iclr/HertzMTAPC23}.
Previous ISR works \cite{yang2023pasd, DBLP:journals/corr/abs-2311-16518} have attempted to utilize existing visual perception models (\ie, ResNet \cite{DBLP:conf/cvpr/HeZRS16}, BLIP \cite{DBLP:conf/icml/0001LXH22}, RAM \cite{DBLP:journals/corr/abs-2306-03514}) to understand image content and provide cross-model semantic priors as prompts.
However, these prompts fall short as they only encompass basic object recognition, lacking high-level information (\eg, spatial positioning and scene understanding) and low-level information (\eg, image quality and sharpness).
Leveraging CLIP \cite{radford2021learning} and additional adapters to align visual inputs with textual inputs into large-scale language models (LLMs, \eg, GPT-4 \cite{achiam2023gpt}, LLaMA \cite{touvron2023llama}), MLLMs \cite{gpt4v, gao2023llama, liu2024visual, zhang2023internlm, zhu2023minigpt, li2023blip} exhibit remarkable visual comprehension capabilities.
By scaling up parameters and training data, MLLMs have shown robust performance in both high-level and low-level perceptual tasks, enabling the extraction of cross-model semantic priors that align with human perception \cite{yin2023survey, wu2023q, you2023depicting}.

\section{Methodology}
\label{sec:method}

\subsection{Cross-modal Semantic Priors}

Current diffusion-based ISR methods typically capture semantic structures from LR images and utilize the T2I model to generate realistic HR images.
However, LR images often undergo complex degradation processes, complicating the task for ISR models to extract semantic content independently.
This difficulty is exacerbated by constraints in scale and the availability of training data, leading to the restoration of blurry and indistinguishable images.
Therefore, it is necessary to provide additional conditions to the diffusion model more effectively.
Considering that T2I models fundamentally generate images guided by textual input, we believe that leveraging textual prompts as semantic priors is a viable approach to enhance the capabilities of T2I models for ISR.

\begin{figure}[t]
  \centering
    \includegraphics[width=0.9\linewidth]{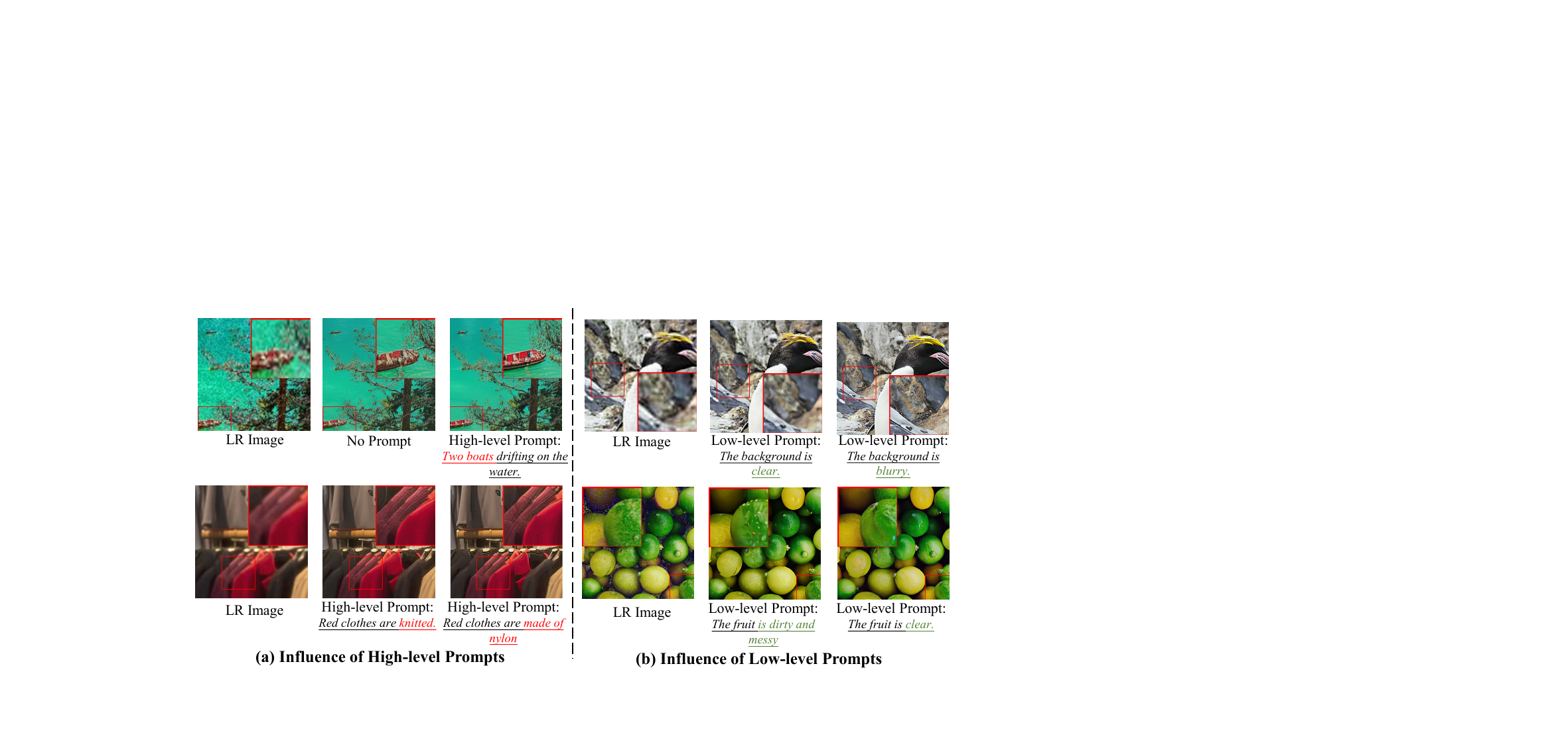}
  \caption{Impact of textual prompts for conditional T2I diffusion models. \textcolor{Red}{\textbf{(a)}} When given high-level prompts containing object categories and detailed textures, high-realism HR images can be restored based on ambiguous LR images. \textcolor{Green}{\textbf{(b)}} When furnished with \textbf{accurate} low-level prompts that encompass distortion types or the general quality of LR images, high-fidelity images can be generated from blurry or noisy inputs.}
  \label{fig:prompt}
\end{figure}

Based on common cognition \cite{freeman2000learning}, the semantic information contained in images can be roughly divided into two categories: \textit{high-level and low-level semantics}.
To more intuitively reflect the impact of different types of prompts on diffusion-based ISR tasks, we have pre-visualized some results in \cref{fig:prompt}.
\textcolor{red}{\textbf{First}}, we provide \textit{high-level semantic priors} related to the LR images, which may encompass information about the objects in the image, spatial positioning, scene descriptions, and more.
As illustrated in \cref{fig:prompt}~(a), it becomes possible to restore images that are clearer and retain the corresponding semantic information (\eg, the boat and the clothes). These show the advantage of utilizing additional high-level semantic priors upon relying on sole diffusion models.
\textcolor{red}{\textbf{Second}}, we use \textit{low-level semantic priors} as guidance, which includes the perception of overall quality, sharpness, noise level, and other distortions about the LR image.
As shown in \cref{fig:prompt}~(b), accurate low-level conditions are necessary for generating high-quality images. 
For the 1st case, where the background of the LR image is highly blurred, providing the correct prompt (\ie, \texttt{"The background is blurry"} ) enables the restoration of a clear image.
Conversely, an incorrect prompt (\ie, \texttt{"The background is clear"}) may lead to the restoration of blurry details due to the model's assumption of sufficient clarity in the LR image.
For the 2nd case, an incorrect prompt (\ie, \texttt{"The fruit is messy and blurry"}) leads to unrealistic artifacts.
This implies that low-level semantic priors assist in modeling the degradation process, which benefits clearer images.
Therefore, we believe that by combining both high- and low-level semantic priors, restoration results can be semantically accurate and rich in detail as well.

\begin{figure}[t]
  \centering
    \includegraphics[width=0.9\linewidth]{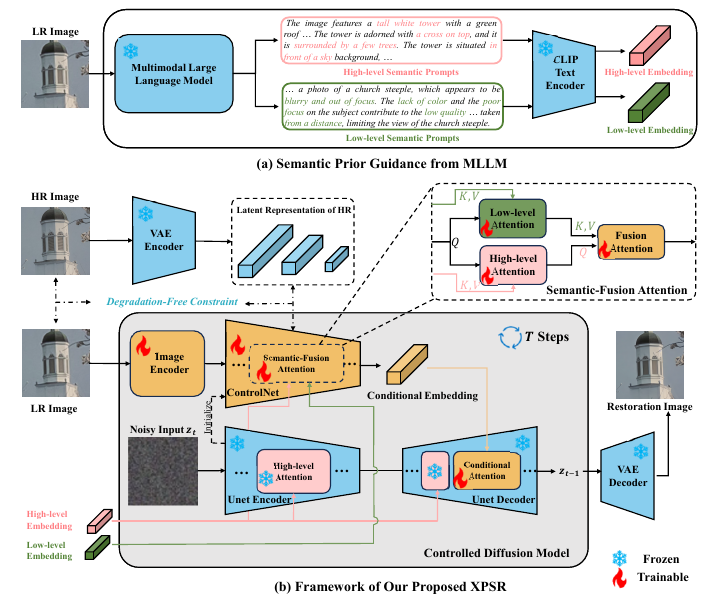}
  \caption{Framework of XPSR. 
  \textcolor{Red}{\textbf{(a)}} First, we integrate an MLLM to acquire semantic priors, encompassing both high-level and low-level descriptions for the LR image. Two varieties of embeddings are derived upon input into the CLIP text encoder.
  \textcolor{Green}{\textbf{(b)}} Next, the LR image, along with the embeddings above, are input into the controlled diffusion model as conditions through a \textit{Semantic-Fusion Attention (SFA)}, adhering to the defined workflow. Besides, a \textit{Degradation-Free Constraint (DFC)} is appended to the ControlNet part, alleviating the challenge of discerning distortions.}
  \label{fig:model}
\end{figure}

\subsection{Framework of XPSR}

To combine semantic priors with the LR image, as conditions of the diffusion model, there still exist three problems that require immediate resolution: \textit{(1) how to obtain semantic priors that describe accurately? (2) How to effectively incorporate these priors with well-defined diffusion models? (3) How to extract semantic-preserved but degradation-unrelated conditions from LR images?}

To address these problems, we propose a framework called \textit{Cross-modal Priors for Super Resolution (XPSR)}. 
As depicted in \cref{fig:model}, it consists of two main stages: generation of semantic priors and image restoration with these priors.
In the first stage (\cref{fig:model}~(a)), \textcolor{Red}{\textbf{to address the first problem}}, we rely on existing SOTA MLLMs to acquire high-level and low-level semantic priors for the LR image (\cref{sec:llava}). Then two varieties of embeddings are derived upon input into the CLIP \cite{radford2021learning} text encoder.
In the second stage (\cref{fig:model}~(b)), we employ the powerful pretrained T2I SD model \cite{rombach2022high} as the backbone.
Additionally, we utilize ControlNet \cite{zhang2023adding} as a controller, guiding the conditional image restoration process.
We clone the encoder of the Unet architecture from SD as a trainable copy to initialize ControlNet.
\textcolor{Red}{\textbf{To address the second problem}}, we design a \textit{Semantic-Fusion Attention (SFA)} module, which facilitates the interaction between the semantic priors obtained in the first stage and the generated priors from the T2I model (\cref{sec:SFA}).
\textcolor{Red}{\textbf{To address the third problem}}, during training, we further attach a \textit{Degradation-Free Constraint (DFC)} to ControlNet, reducing the impact of degradation and extracting semantic-aware representations from the LR image (\cref{sec:DFC}). During inference, only ControlNet and Unet parts, along with the MLLM, are used for predicting restoration images.

\subsection{Semantic Prompts from MLLM}
\label{sec:llava}


By scaling up the size of data and the scale of models, MLLMs have demonstrated impressive capabilities in semantic comprehension \cite{yin2023survey}.
Accordingly, we employ a noteworthy MLLM called LLaVA \cite{liu2024visual} to perceive the LR images and extract semantic priors. 
By defining appropriate instructions, we aim to facilitate its understanding of semantics from both high- and low-level perspectives.

\begin{figure}[t]
  \centering
    \includegraphics[width=0.86\linewidth]{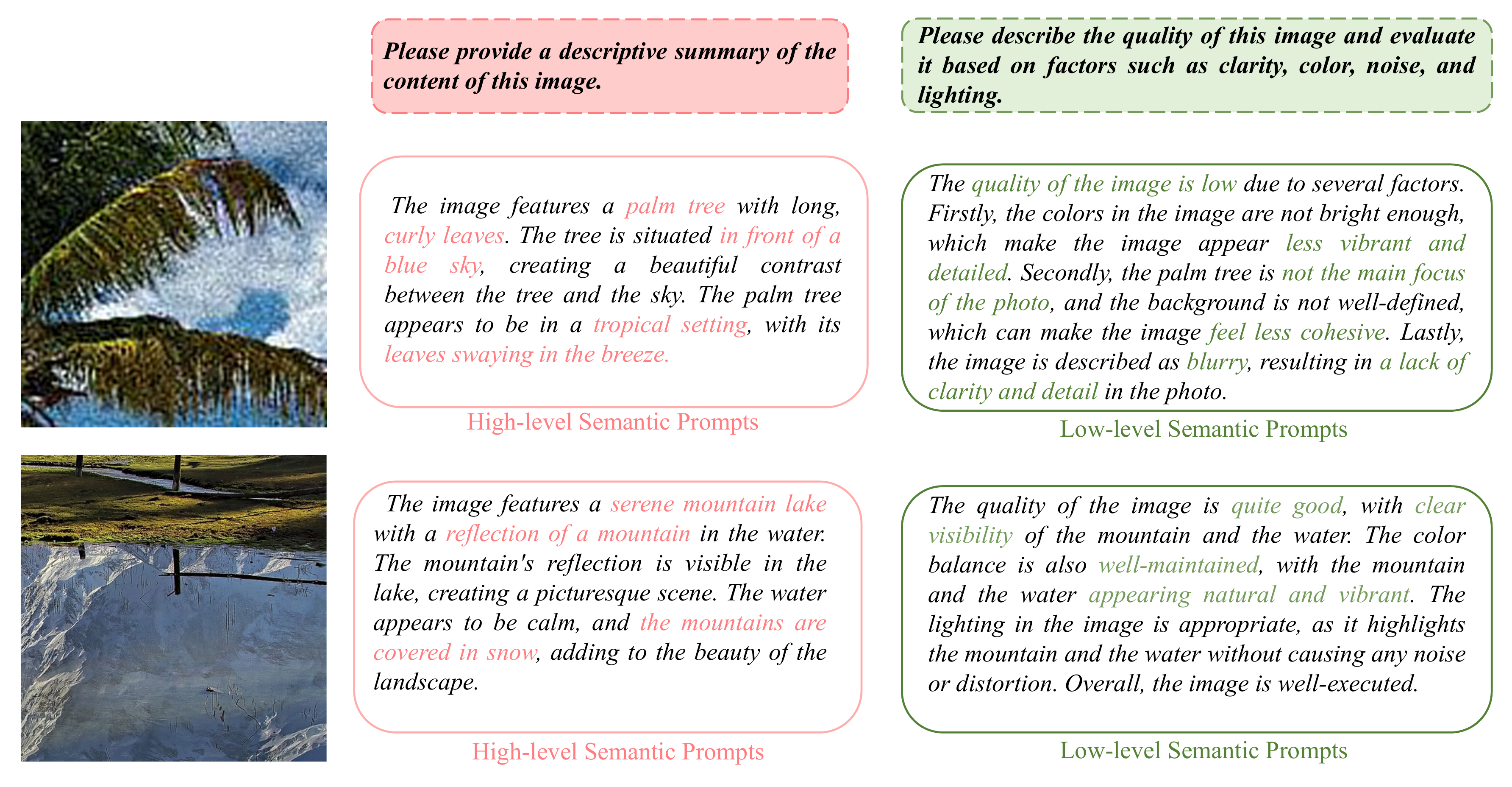}
  \caption{Given appropriate instructions, LLaVA can generate high- and low-level semantic prompts consistent with human perception for both high- and low-quality images.}
  \label{fig:llava}
\end{figure}

Determining appropriate instructions to guide the MLLMs in generating outputs that align with human preferences is a matter that requires iterative practice \cite{cheng2023black}.
\textit{In the acquisition of high-level semantic priors}, we have ultimately opted for the instruction \texttt{"Please provide a descriptive summary of the content of this image"} as the guidance for LLaVA.
In our practical experience, we observe that similar instructions consistently yield similar results with LLaVA, generating the intended high-level semantics, including object descriptions, spatial locations, scenes, and other relevant content.
\textit{In the acquisition of low-level semantic priors}, we first analyze the distribution of degradations in real-world scenarios.
The degradation primarily occurs due to processes such as capturing, transcoding compression, or transmission. 
Consequently, it introduces various distortions such as motion blur, low-light conditions, color shifts, noise, and compression artifacts \cite{DBLP:journals/tip/MittalMB12, DBLP:journals/tip/ZhangZB15, DBLP:journals/tip/SheikhSB06, DBLP:journals/tip/HosuLSS20,DBLP:journals/tip/HosuLSS20,DBLP:conf/mm/LiuWYSTZWL23,DBLP:conf/mm/YuanKZSW23,DBLP:journals/corr/abs-2303-00521}.
Therefore, by incorporating guidance in the instructions on these quality-related factors, we observe that this approach generates more realistic and detailed descriptions.
We adopt the instruction \texttt{"Please describe the quality of this image and evaluate it based on factors such as clarity, color, noise, and lighting"} to generate descriptions related to overall quality, clarity, noise level, color, and other relevant information.
As shown in Fig. \ref{fig:llava}, LLaVA is capable of providing high-level and low-level semantic descriptions that align with human perception.

\subsection{Semantic Priors Fusion for Diffusion Model}
\label{sec:SFA}

To integrate extra semantic priors into the diffusion model, an intuitive and naive approach is to employ two consecutive cross-attention (\ie, attention type for the original T2I) structures as a serial connection into the ControlNet. 
Given the feature $\mathbf{x}_k$ in the $k$-th layer, the computation process can be noted as:
\begin{equation}
    \begin{split}
    \mathbf{x}_{k+1} = \mathcal{CA}_l(\mathcal{CA}_h(\mathbf{x}_k, {c}_h), {c}_l),
    \end{split}
\end{equation}
where $\mathcal{CA}_h$ and $\mathcal{CA}_l$ represent the high-level and low-level cross-attention while ${c}_h$ and ${c}_l$ denote the two types of semantic prompts, respectively.
However, as these two types of priors are distinct, serial processing may cause a certain part of the information to be overwritten and obtain suboptimal results.

Therefore, we design a new \textit{Semantic-Fusion Attention (SFA)}, which employs two parallel branches of cross-attention, and then fuses them to obtain the final condition using a fusion attention $\mathcal{CA}_f$. As shown in \cref{fig:model}, the fusion attention obtains the query $Q$ from the high-level results, and the key $K$ and value $V$ from the low-level results. The parallel type of SFA can be written as:
\begin{equation}
    \begin{split}
    \mathbf{x}_{k+1} = \mathcal{CA}_f(\mathcal{CA}_h(\mathbf{x}_k, {c}_h), \mathcal{CA}_l(\mathbf{x}_k, {c}_l)).
    \end{split}
\end{equation}
In this way, SFA can achieve a balance between priors from different levels, allowing for an adaptive selection of semantic features. Ablation in \cref{tab:DFC} further verifies the effectiveness of this setting.
Specifically, we claim that a low-level understanding of LR images is not necessary for Unet, which receives noise as input, so only high-level attention is used.
To enhance the generalizability of SD for downstream tasks and fortify the fusion of features derived from LR images, we add a cross-attention module termed \textit{Conditional Attention} in the Unet.

\subsection{Degradation-Free Constraint}
\label{sec:DFC}


Real-world images may experience a variety of degradations, such as blurriness, blocky artifacts, \etc, leading to distortions that affect both high-frequency and low-frequency components (appeared in the pixel- and latent-space) \cite{DBLP:journals/tip/SheikhSB06, DBLP:journals/tip/HosuLSS20}.
Aiming to mitigate the impact of degradations and extract robust semantic information from LR images, we propose a \textit{Degradation-Free Constraint (DFC)} which attaches constraints on both \textit{the pixel space} and \textit{the latent space}.

As shown in \cref{fig:model}, the LR image is initially passed through a pyramid-shaped image encoder in ControlNet for downsampling. 
Following \cite{yang2023pasd}, we apply a pixel-space constraint in the image encoder. At the $i$-th layer of the pyramid encoder, we employ a single-layer convolution to map the feature map into an image $\hat{x}_i \in \mathbb{R}^{H_i\times W_i \times 3}$ with RGB channels.
As the scale is reduced by half compared to the previous layer, we apply an $L_1$ loss to make $\hat{x}_i$ as close as possible to the downsampled result of the original HR image $x_{\textit{hr,i}}$, denoted as $\lVert x_{\textit{hr,i}} - \hat{x}_i \Vert_1$.

Besides, ControlNet adopts a pyramid-like Unet Encoder to encode semantic features at the latent level, we further apply a similar latent-space constraint.
The feature map obtained from the $j$-th layer is mapped to the latent space as $\hat{z}_j$.
We align it with the downsampled result of the HR latent $z_{\textit{hr,j}}$ at scale $j$ using the $L_1$ loss 
 $\lVert z_{\textit{hr,j}} - \hat{z}_j \Vert_1$.
The final DFC is a combination of constraints in both the pixel space and the latent space, which can be noted as:
\begin{equation}
    \begin{split}
    \mathcal{L}_{DFC} = \sum\nolimits_{i=1}^3 \lVert x_{\textit{hr,i}} - \hat{x}_i \Vert_1 + \sum\nolimits_{j=1}^3 \lVert z_{\textit{hr,j}} - \hat{z}_j \Vert_1.
    \end{split}
\end{equation}

\subsection{Training and Testing Strategy}
\label{sec:3.6}

During training, the HR image is mapped to the latent embedding $z_{\textit{hr}}$.
We add noise $\epsilon$ through the diffusion process by following $t$ steps to obtain the noisy latent $z_{\textit{hr}}^t$, where $t$ is randomly sampled from a range of $[1, T]$.
Overall, XPSR relies on the LR image $x_{\textit{lr}}$, noisy latent $z_{\textit{hr}}^t$, high-level prompt $c_h$, and low-level prompt $c_l$ to predict the added noise. The optimization objective is:
\begin{equation}
    \begin{split}
    \mathcal{L}_{D}=\mathbb{E}_{z_{\textit{hr}}, t, c_h, c_l, x_{\textit{lr}}, \epsilon \sim \mathcal{N}(0,1)}\left[\left\|\epsilon-\epsilon_\theta\left(z_{\textit{hr}}^t, t, c_h, c_l,  x_{\textit{lr}}\right)\right\|_2^2\right] .
    \end{split}
\end{equation}
where $\epsilon_\theta(\cdot)$ represents the mapping function of XPSR. The final loss function is a weighted sum of the diffusion loss and the DFC loss:
\begin{equation}\label{equ:loss}
    \begin{split}
    \mathcal{L}= \mathcal{L}_{D} + \lambda \mathcal{L}_{DFC},
    \end{split}
\end{equation}
where $\lambda$ is a balancing coefficient.
To reduce training costs and leverage the generative prior of SD, we freeze all the parameters of SD during training and only train the ControlNet and the added \textit{Conditional Attention}.

During testing, we employ the classifier-free guidance strategy \cite{ho2022classifier}, which allows the diffusion model to generate better-quality images through negative prompts without additional training.
In each step, we rely on regular high-level prompts $c_h$ for prediction while simultaneously replacing $c_h$ with negative prompts $c_{neg}$. 
These predictions are fused to obtain the final output:
\begin{equation}\label{equ:negative}
    \begin{split}
    &\hat{\epsilon} = \epsilon_\theta(z^t_{\textit{lr}}, t, c_h, c_l,  x_{\textit{lr}})\\
    &\hat{\epsilon}_{neg} = \epsilon_\theta(z^t_{\textit{lr}}, t, c_{neg}, c_l,  x_{\textit{lr}})\\
    &\widetilde{\epsilon} =\hat{\epsilon} + \lambda_s(\hat{\epsilon} -\hat{\epsilon}_{neg})
    \end{split}
\end{equation}
where $\lambda_s$ is the guidance scale and $z^t_{\textit{lr}}$ is the noisy latent of LR image. In practice, We employ combinations of negative words "\textit{blurry, dotted, noise, unclear, low-res, over-smoothed}" as negative prompts to generate higher-quality images.


\section{Experiments}

\subsection{Experimental Settings}

\paragraph{Training and testing datasets.}
We train our XPSR on DIV2K \cite{DBLP:conf/cvpr/AgustssonT17}, DIV8K \cite{DBLP:conf/iccvw/GuLDFLT19}, Flickr2K \cite{timofte2017ntire}, Unsplash2K \cite{kim2021noise}, OST \cite{DBLP:conf/cvpr/WangYDL18}, and the first 5K face images from FFHQ \cite{DBLP:journals/pami/KarrasLA21}. 
We use the degradation pipeline of Real-ESRGAN \cite{DBLP:conf/iccvw/WangXDS21} to synthesize LR-HR image pairs for training.
To conduct a comprehensive evaluation of ISR, we conduct testing using synthetic and real-world datasets. 
The synthetic dataset is generated from the DIV2K validation set, where we randomly crop 3K patches following the same degradation pipeline.
Following \cite{DBLP:journals/corr/abs-2305-07015,DBLP:journals/corr/abs-2311-16518}, the LR image are center-cropped to $128\times128$.
For the real-world dataset, we employ the DrealSR and RealSR datasets for evaluation, with each image being center-cropped.
Each HR image in the training and testing sets has a resolution of $512\times512$.

\paragraph{Implementation details.}
We employ the pretrained SD-v1.5\footnote{https://huggingface.co/runwayml/stable-diffusion-v1-5} as the base T2I model.
For training, we utilize an AdamW \cite{loshchilov2017fixing} optimizer with a weight decay of 1e-2 to finetune XPSR for 100K iterations. 
The batch size and the learning rate are set to 32 and 5e-5. 
All experiments are conducted on 8 NVIDIA A800 GPUs.
For inference, we adopt DDPM sampling \cite{DBLP:conf/icml/NicholD21} with 20 timesteps. The balancing coefficient $\lambda$ in \cref{equ:loss} is 0.05 and the guidance scale $\lambda_s$ in \cref{equ:negative} is 5.5.

\paragraph{Evaluation metrics.}
We employ a range of widely used reference and non-reference metrics to conduct a comprehensive quantitative evaluation of ISR methods. 
In reference-based metrics, PSNR and SSIM \cite{DBLP:journals/tip/WangBSS04} (calculated on the Y channel in YCbCr space) are fidelity metrics, while LPIPS \cite{DBLP:conf/cvpr/ZhangIESW18}, DISTS \cite{DBLP:journals/pami/DingMWS22} are quality evaluation metrics.
FID \cite{DBLP:conf/nips/HeuselRUNH17} calculates the distance between the distributions of generated images and reference images. 
MANIQA \cite{DBLP:conf/cvpr/YangWSLGCWY22}, MUSIQ \cite{ke2021musiq}, and CLIPIQA \cite{wang2023exploring} are
non-reference image quality assessment (IQA) metrics.

\subsection{Comparisons with SOTA Methods}
\begin{figure}[t]
  \centering
    \includegraphics[width=0.78\linewidth]{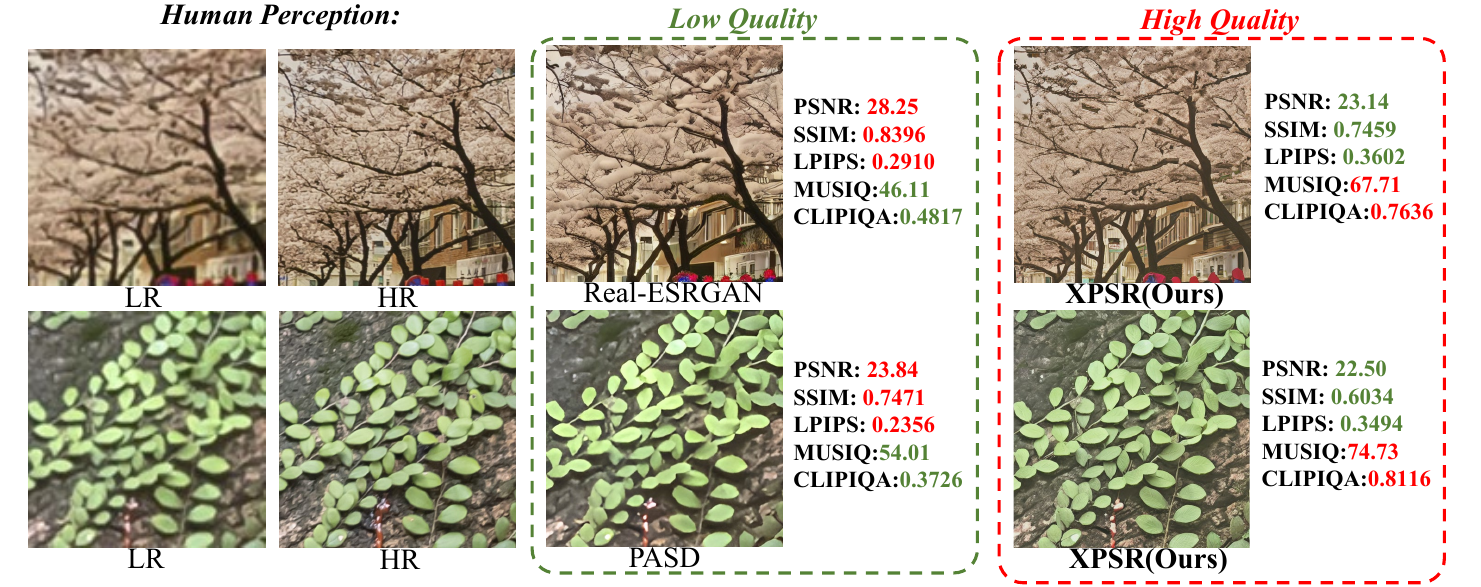}
  \caption{Drawbacks of current full-reference metrics (\eg, PSNR, SSIM, LPIPS). The visualization results show that XPSR generates images with high-realism and high-fidelity in human perception, but obtains lower scores in some cases.}
  \label{fig:metrics}
\end{figure}
\begin{table}[t]
  \centering
  \tiny
  \caption{Quantitative comparison with SOTA methods on both synthetic and real-world benchmarks. 
  \textcolor{red}{Red} and \textcolor{blue}{blue} colors are the best and second-best performance.
  }
  \label{tab:headings}
  \centering
  \begin{tabular}{c|c|ccc|ccccc|c}
    \toprule
    \multirow{2}{*}{Dataset} & \multirow{2}{*}{Metrics} & \multicolumn{3}{c|}{GAN-based SR} & \multicolumn{6}{c}{Diffusion-based SR} \\
    & & BSRGAN & Real-ESRGAN & SwinIR & LDM & StableSR & DiffBIR & PASD & SeeSR & XPSR \\
    \midrule
    \multirow{9}{*}{\textit{DIV2K-Val}}
    & PSNR$\uparrow$     & \textcolor{red}{24.42}& \textcolor{blue}{24.30}& 23.77& 21.66& 23.26& 23.49& 23.59& 23.56& 22.80 \\
    & SSIM$\uparrow$     & 0.6164& \textcolor{red}{0.6324}& \textcolor{blue}{0.6186}& 0.4752& 0.5670& 0.5568& 0.5899& 0.5981&0.5627\\
    & LPIPS$\downarrow$  & 0.3511&\textcolor{blue}{0.3267}& 0.3910& 0.4887& \textcolor{red}{0.3228}& 0.3638& 0.3611& 0.3283&0.3761 \\
    & DISTS$\downarrow$  & 0.2369& 0.2245& 0.2291& 0.2693& \textcolor{blue}{0.2116}& 0.2177& 0.2134&\textcolor{red}{0.2008}&0.2217 \\
    & FID$\downarrow$    & 50.99& 44.34& 44.45& 55.04& \textcolor{red}{28.32}& 34.55& 39.74& \textcolor{blue}{28.89}&33.38 \\
    & MANIQA$\uparrow$   & 0.3547& 0.3756& 0.3411& 0.3589& 0.4173& 0.4598& 0.4440& \textcolor{blue}{0.5046}&\textcolor{red}{0.6080} \\
    & CLIPIQA$\uparrow$  & 0.5253& 0.5205& 0.5213& 0.5570& 0.6752& 0.6731&0.6573& \textcolor{blue}{0.6959}&\textcolor{red}{0.7816} \\
    & MUSIQ$\uparrow$    & 60.18& 59.76& 57.21& 57.46& 65.19& 65.57& 66.58& \textcolor{blue}{68.35}&\textcolor{red}{69.99} \\

    \midrule
    \multirow{9}{*}{\textit{RealSR}}
    & PSNR$\uparrow$     & \textcolor{red}{26.38}& 25.68& \textcolor{blue}{25.88}& 25.66& 24.69& 24.94& 25.21& 25.31&24.19 \\
    & SSIM$\uparrow$     & \textcolor{blue}{0.7651}& 0.7614& \textcolor{red}{0.7671}& 0.6934& 0.7090& 0.6664& 0.7140& 0.7284&0.6870 \\
    & LPIPS$\downarrow$  & \textcolor{blue}{0.2656}& 0.2710& \textcolor{red}{0.2614}& 0.3367& 0.3003& 0.3485& 0.2986& 0.2993&0.3517 \\
    & DISTS$\downarrow$  & 0.2124& \textcolor{red}{0.2060}& \textcolor{blue}{0.2061}& 0.2324& 0.2134& 0.2257& 0.2125& 0.2224&0.2471 \\
    & FID$\downarrow$    & 141.25& 135.14& 132.80& 133.34& 131.72& \textcolor{blue}{127.59}& 139.42& \textcolor{red}{126.21}& 141.95 \\
    & MANIQA$\uparrow$   & 0.3763& 0.3736& 0.3561& 0.3375& 0.4167& 0.4378& 0.4418& \textcolor{blue}{0.5370}& \textcolor{red}{0.6059} \\
    & CLIPIQA$\uparrow$  & 0.5114& 0.4487& 0.4433& 0.6053& 0.6200& 0.6396& 0.6009& \textcolor{blue}{0.6638}& \textcolor{red}{0.7354} \\
    & MUSIQ$\uparrow$    & 63.28& 60.37& 59.28& 56.32& 65.25& 64.32& 66.61&\textcolor{blue}{ 69.56}& \textcolor{red}{70.23} \\

    \midrule
    \multirow{9}{*}{\textit{DRealSR}}
    & PSNR$\uparrow$     & \textcolor{red}{28.70} & \textcolor{blue}{28.61}& 28.20& 27.78& 27.87& 26.57& 27.45& 28.13& 26.62 \\
    & SSIM$\uparrow$     & \textcolor{blue}{0.8028}& \textcolor{red}{0.8052}& 0.7983& 0.7152&  0.7427& 0.6516& 0.7539& 0.7711& 0.7220 \\
    & LPIPS$\downarrow$  & 0.2858& \textcolor{red}{0.2819}& \textcolor{blue}{0.2830}& 0.3745& 0.3333& 0.4537& 0.3331& 0.3142&0.3864 \\
    & DISTS$\downarrow$  & 0.2144& \textcolor{red}{0.2089}& \textcolor{blue}{0.2103}& 0.2417& 0.2297& 0.2724& 0.2322& 0.2230&0.2606 \\
    & FID$\downarrow$    & 155.62& 147.66& \textcolor{red}{146.38}& 164.87& 148.18& 160.67& 173.40& \textcolor{blue}{147.00}&164.68 \\
    & MANIQA$\uparrow$   & 0.3441& 0.3435& 0.3311& 0.3342& 0.3897&0.4602& 0.4551& \textcolor{blue}{0.5077}& \textcolor{red}{0.5713} \\
    & CLIPIQA$\uparrow$  & 0.5061& 0.4525& 0.4522& 0.5984& 0.6321& 0.6445& 0.6365&\textcolor{blue}{0.6893} & \textcolor{red}{0.7360} \\
    & MUSIQ$\uparrow$    & 57.16& 54.27& 53.01& 51.37& 58.72&61.06& 63.69& \textcolor{blue}{64.75} & \textcolor{red}{67.84} \\
  \bottomrule
  \end{tabular}
\end{table}

To verify the effectiveness, we compare XPSR with other SOTA GAN-based and Diffusion-based ISR methods\footnote{All methods are tested based on their official code and models.}, \ie, BSRGAN~\cite{DBLP:conf/iccv/0008LGT21}, Real-ESRGAN~\cite{DBLP:conf/iccvw/WangXDS21}, SwinIR-GAN~\cite{DBLP:conf/iccvw/LiangCSZGT21}, LDM~\cite{DBLP:conf/cvpr/RombachBLEO22}, StableSR~\cite{DBLP:journals/corr/abs-2305-07015}, DiffBIR~\cite{DBLP:journals/corr/abs-2308-15070}, PASD~\cite{yang2023pasd} and SeeSR \cite{DBLP:journals/corr/abs-2311-16518}.

\paragraph{Quantitative comparisons.}

In \cref{tab:headings}, some observations can be found. \textcolor{Red}{\textbf{First}}, XPSR outperforms other SOTA methods in MAINIQA, CLIPIQA, and MUSIQ across all datasets by large margins, reflecting excellent restoration results. For example, XPSR surpasses the second-best method SeeSR by \textbf{10.34\%, 8.57\%, and 1.64\%} in DIV2K-val, respectively. \textcolor{Red}{\textbf{Second}}, Diffusion-based methods generally fall behind GAN-based methods in reference metrics (\eg, PSNR, SSIM). This is mainly because Diffusion-based methods can produce more realistic details but at the expense of fidelity to the LR images. Furthermore, we give several examples that highlight the limitations of current reference metrics in \cref{fig:metrics}. Our restoration results exhibit \textbf{higher quality in human perception, yet they lag in reference metrics.} This phenomenon has also been confirmed in many previous studies \cite{DBLP:journals/corr/abs-2305-07015,yang2023pasd,DBLP:journals/corr/abs-2311-16518}. We believe there is a need to update the metrics used to evaluate ISR, aiming to closely mimic human perception.

\paragraph{Qualitative comparisons.}


\begin{figure}[t]
  \centering
    \includegraphics[width=\linewidth]{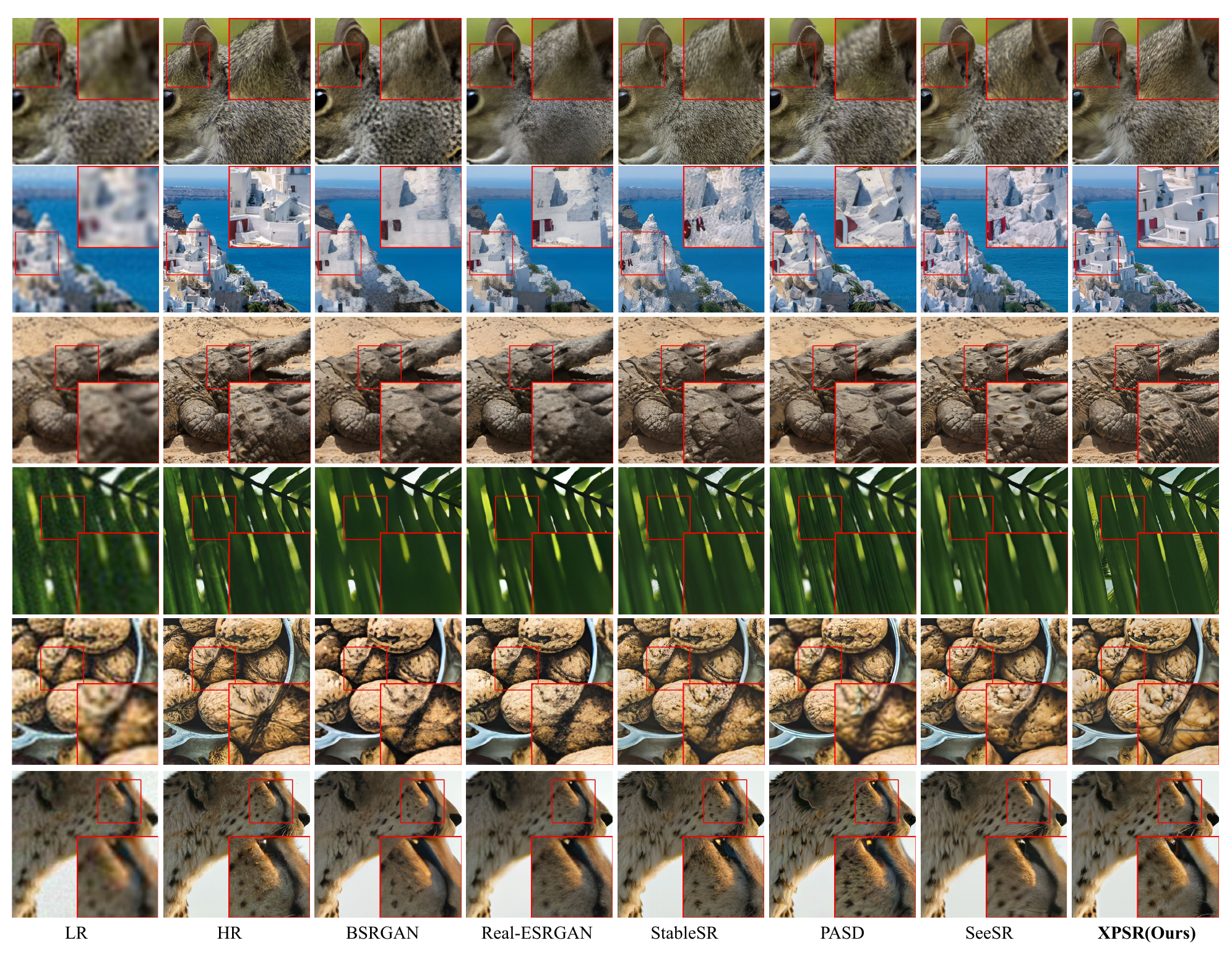}
  \caption{Qualitative comparisons with different SOTA methods. XPSR is adept at precisely restoring the textures and details of specific objects, such as hair and buildings, even under challenging conditions of degradation. \textbf{Zoom in for a better view}.}
  \label{fig:sota}
\end{figure}

In \cref{fig:sota}, we present visual comparisons on the test set, revealing key observations.
\textcolor{Green}{\textbf{First}}, diffusion-based methods can generate more realistic results than GAN-based methods, showing the advantage of inherent \textit{generative priors}.
\textcolor{Green}{\textbf{Second}}, XPSR surpasses StableSR in generating results that are semantically accurate and rich in detail (\eg, the 1st, 2nd, 4th, and 5th row), exhibiting the benefit of employing high-level semantic conditions.
\textcolor{Green}{\textbf{Third}}, XPSR also outperforms PASD and SeeSR in detail generation and degradation removal. In the 2nd row, XPSR distinguishes itself as \textbf{the sole method} adept at understanding the semantic context of the LR image, thus producing detailed and lucid depictions of the house. Moreover, in the 1st, 4th, and 5th rows, XPSR can produce more detailed animal hair, clearer leaves, and walnut textures, respectively. 
These examples highlight the benefits of utilizing \textit{cross-modal semantic priors} from MLLM over-relying on dispersed object tagging information.
We are equally impressed by the generative capabilities of all diffusion-based methods, as the outcomes in row 6 show their tremendous potential in the realm of ISR.

\subsection{Ablation Study}

\begin{figure}[t]
  \centering
    \includegraphics[width=0.9\linewidth]{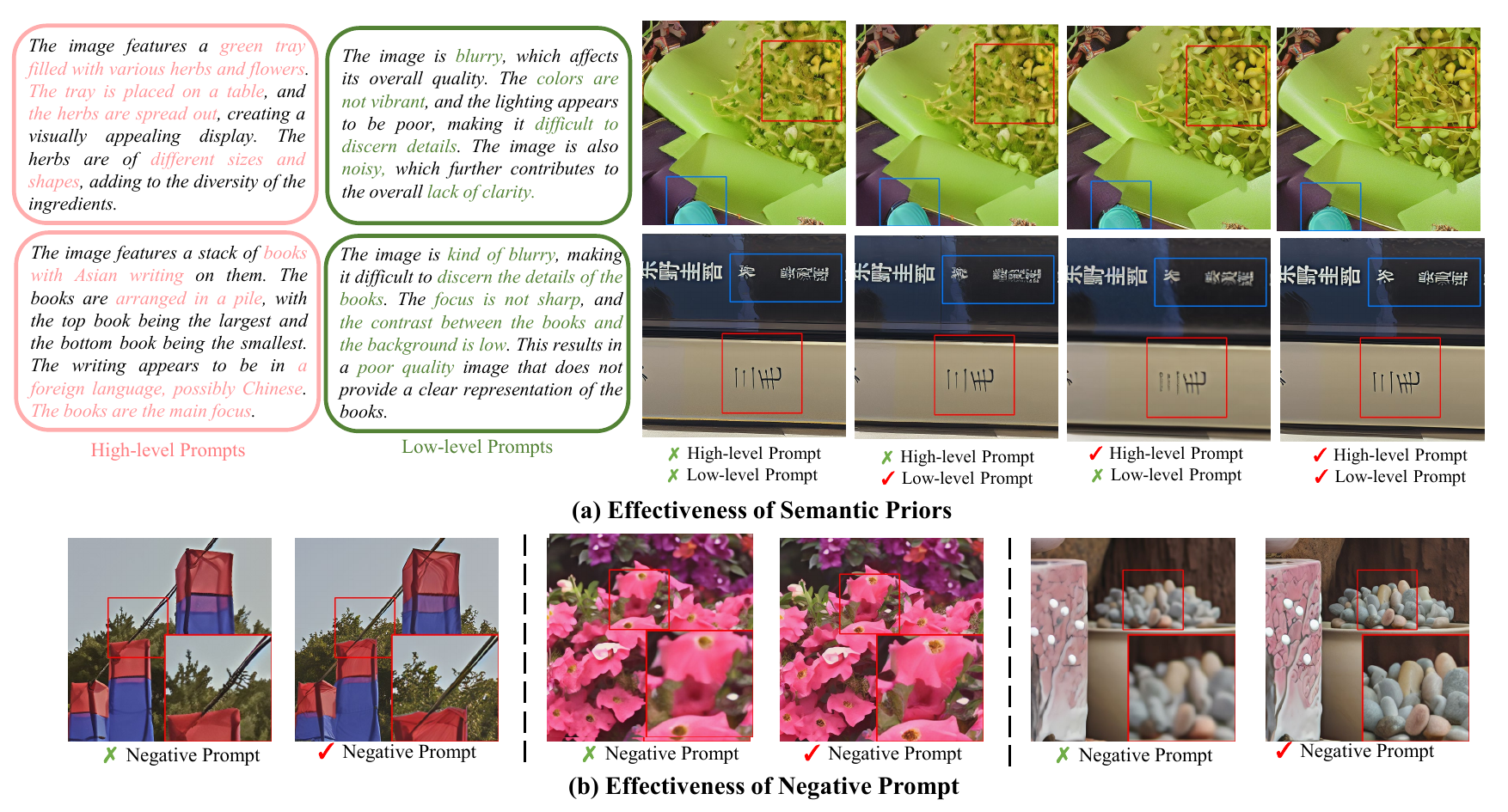}
  \caption{Effectiveness of cross-modal semantic priors and negative prompts.}
  \label{fig:prompt+negative}
\end{figure}
\begin{table}[t]
  \centering
  \tiny
  \caption{Ablation on the types of semantic priors.}
  \label{tab:semantic}
   \centering
  \begin{tabular}{cc|ccccc|ccccc}
    \toprule
    \multicolumn{2}{c|}{Semantic Priors} & \multicolumn{5}{c|}{\textit{DrealSR}} & \multicolumn{5}{c}{\textit{RealSR}} \\
    high-level & low-level & SSIM$\uparrow$ &LPIPS$\downarrow$& FID$\downarrow$ & MANIQA$\uparrow$ & MUSIQ$\uparrow$
    & SSIM$\uparrow$  &LPIPS$\downarrow$&FID$\downarrow$ & MANIQA$\uparrow$ & MUSIQ$\uparrow$ \\
    \midrule
    \XSolidBrush & \XSolidBrush & 0.7157 & 0.4033 & 188.78 & 0.6078 & 67.31 & 0.6767 & 0.3739 & 157.73 & 0.6426 & 70.95\\
    \XSolidBrush & \Checkmark   & 0.7114 & 0.4145 & 195.37 & 0.6213 & 68.73 & 0.6710 & 0.3796 & 162.68 & 0.6438 & 71.05\\
    \Checkmark & \XSolidBrush   & 0.7254 & 0.3776 & 160.78 & 0.5505 & 65.79 & 0.6938 & 0.3461 & 139.54 & 0.5752 & 68.74\\
    \midrule
    \Checkmark & \Checkmark     & 0.7220& 0.3864 & 164.68 & 0.5713 & 67.84 & 0.6870 & 0.3517 & 141.95 & 0.6059 & 70.23\\
  \bottomrule
  \end{tabular}
\end{table}

\paragraph{Effectiveness of semantic priors.}

To assess the influence of cross-modal semantic priors, the trained XPSR is utilized for inference across various input conditions. The results are given in \cref{tab:semantic} and \cref{fig:prompt+negative}. 
\textcolor{Red}{\textbf{First}}, the absence of high-level priors leads to a substantial decrease in fidelity metrics (\ie, SSIM, FID). This can be explained that the loss of high-level semantics promotes SFA to overly rely on low-level priors for degradation removal. Although this may yield some gains to MANIQA, it will lead to discrepancies in the content of the restoration images as given in \cref{fig:prompt+negative}~(a).
\textcolor{Red}{\textbf{Second}}, the absence of low-level priors significantly worsens the quality metrics (\ie, MANIQA, MUSIQ). This occurs as SFA gives precedence to high-level priors, leading to a marginal enhancement in SSIM, yet it may produce images that are excessively blurry or marred by noise and artifacts. 
\textcolor{Red}{\textbf{Overall}}, the integration of dual-level priors facilitates balancing, permitting the restorations that preserve semantic details while also showcasing vivid content.

\begin{figure}[t]
  \centering
    \includegraphics[width=0.9\linewidth]{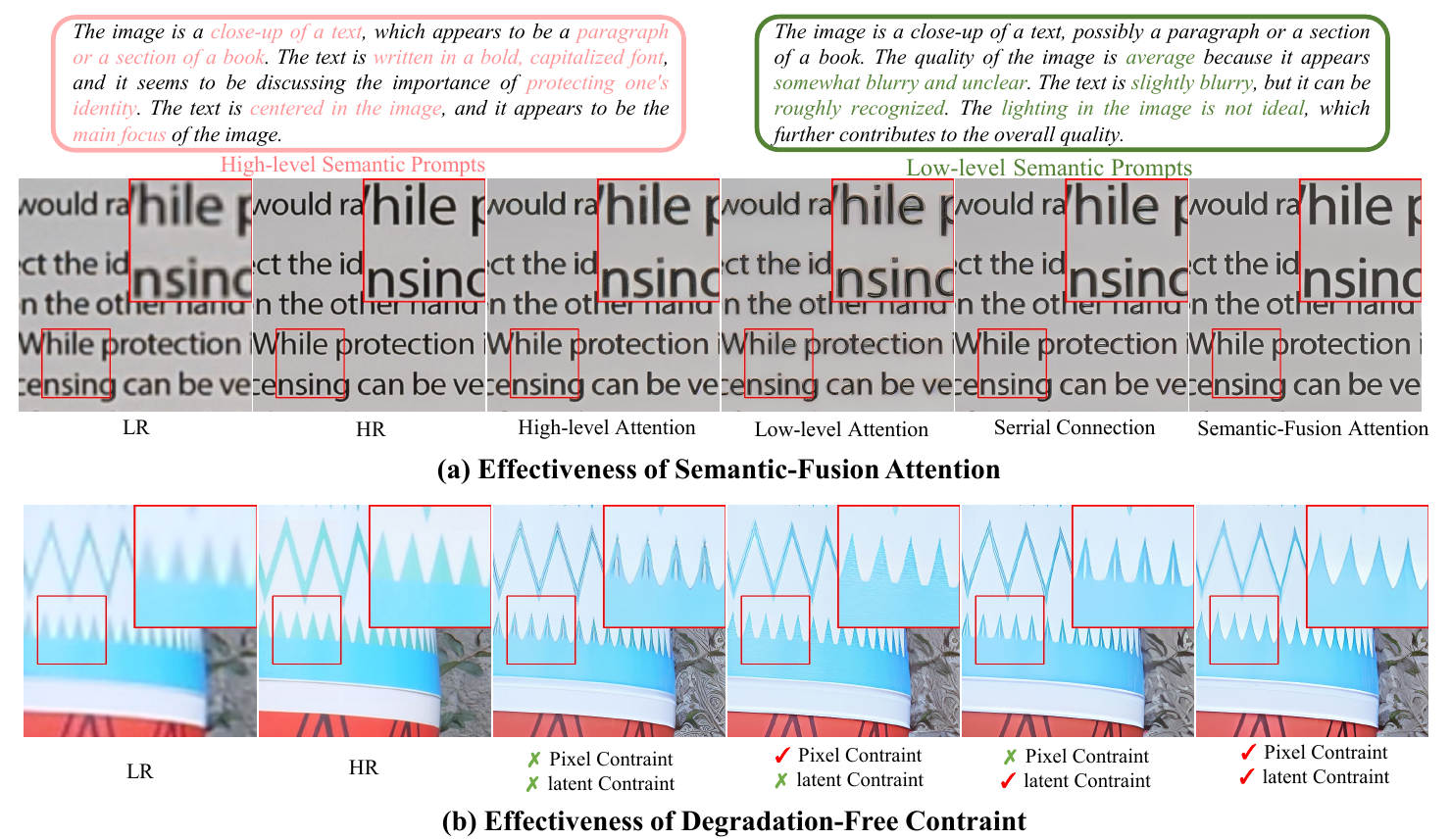}
  \caption{Effectiveness of \textit{Semantic-Fusion Attention} and \textit{Degradation-Free Constraint}.}
  \label{fig:sfa+dfc}
\end{figure}

\begin{table}[t]
  \centering
  \tiny
  \caption{Ablation on the SFA using different fusion types.}
  \label{tab:fusion}
   \centering
  \begin{tabular}{c|ccccc|ccccc}
    \toprule
    \multirow{2}{*}{\makecell{Semantic-Fusion \\ Methods}} & \multicolumn{5}{c|}{\textit{DrealSR}} & \multicolumn{5}{c}{\textit{RealSR}} \\
    ~ & SSIM$\uparrow$ &LPIPS$\downarrow$& FID$\downarrow$ & MANIQA$\uparrow$ & MUSIQ$\uparrow$
    & SSIM$\uparrow$  &LPIPS$\downarrow$&FID$\downarrow$ & MANIQA$\uparrow$ & MUSIQ$\uparrow$ \\
    \midrule
    High-level& 0.7298&0.3674&160.04&0.5472&65.04&0.6878&0.3500&134.68&0.5911&68.89\\
    Low-level& 0.7578&0.3616&161.04&0.5034&63.97&0.7047&0.3465&144.71&0.5305&66.87\\
    Serial connection & 0.6834& 0.4134 & 155.23 & 0.5614 & 65.99 & 0.6526 & 0.3908 & 137.49 & 0.6034 & 69.69\\
    \midrule
    \textbf{SFA(Ours)} & 0.7220& 0.3864 & 164.68 & 0.5713 & 67.84 & 0.6870 & 0.3517 & 141.95 & 0.6059 & 70.23\\
  \bottomrule
  \end{tabular}
\end{table}

\paragraph{Effectiveness of Semantic-Fusion Attention.}

Besides the adopted parallel cross-attention mechanisms in SFA, we further test using: (1) only high-level cross-attention, (2) only low-level cross-attention, and (3) a serial concatenation cross-attention. As shown in \cref{tab:fusion} and \cref{fig:sfa+dfc}~(a), the parallel version achieves the highest results (\ie, generating clear text without any artifacts).
\textcolor{Green}{\textbf{First}}, it further verifies the effectiveness of jointly utilizing high-level and low-level semantic priors to enhance ISR.
\textcolor{Green}{\textbf{Second}}, this confirms the efficacy of employing multiple types of conditions in parallel as opposed to a straightforward serial approach.

\paragraph{Effectiveness of Degradation-Free Constraint.}

\begin{table}[t]
  \centering
  \tiny
  \caption{Ablation on the \textit{Degradation-Free Constraint}.}
  \label{tab:DFC}
   \centering
  \begin{tabular}{cc|ccccc|ccccc}
    \toprule
    \multicolumn{2}{c|}{\makecell{Degradation-Free \\ Constraint}} &\multicolumn{5}{c|}{\textit{DrealSR}} & \multicolumn{5}{c}{\textit{RealSR}} \\
    Pixel &Latent & SSIM$\uparrow$ &LPIPS$\downarrow$& DISTS$\downarrow$ & MANIQA$\uparrow$ & MUSIQ$\uparrow$
    & SSIM$\uparrow$  &LPIPS$\downarrow$&DISTS$\downarrow$ & MANIQA$\uparrow$ & MUSIQ$\uparrow$ \\
    \midrule
    \XSolidBrush &\XSolidBrush & 0.6979& 0.3943&0.2649&0.5684&67.50&0.6341&0.3677&0.2525&0.6043&70.28\\
    \Checkmark &\XSolidBrush & 0.7095&0.3938& 0.2610&0.5445&64.29& 0.6783&0.3629&0.2610&0.6042&69.91\\
    \XSolidBrush &\Checkmark &0.7066&0.3848 & 0.2580&0.5774&67.62&0.6681 & 0.3665& 0.2537 & 0.6000 & 70.08\\
    \midrule
    \Checkmark &\Checkmark  & 0.7220& 0.3864 & 0.2606 & 0.5713 & 67.84 & 0.6870 & 0.3517 & 0.2471 & 0.6059 & 70.23\\
  \bottomrule
  \end{tabular}
\end{table}

\begin{table}[t]
  \centering
  \tiny
  \caption{Ablation on the negative prompts and the base T2I diffusion models.}
   \centering
\begin{subtable}{\textwidth}
    \centering
   \begin{tabular}{c|ccccc|ccccc}
    \toprule
    \multirow{2}{*}{Negative Prompts}  &\multicolumn{5}{c|}{\textit{DrealSR}} & \multicolumn{5}{c}{\textit{RealSR}} \\
    & SSIM$\uparrow$ &LPIPS$\downarrow$& FID$\downarrow$ & MANIQA$\uparrow$ & MUSIQ$\uparrow$
    & SSIM$\uparrow$  &LPIPS$\downarrow$&FID$\downarrow$ & MANIQA$\uparrow$ & MUSIQ$\uparrow$ \\
    \midrule
    \XSolidBrush &0.7355&0.3731 &157.32 &0.4778&61.70 & 0.6987&0.3419 &134.22 &0.5241&65.89 \\
    \Checkmark  &0.7220&0.3864 &164.68 & 0.5713&67.84&0.6870 &0.3517& 141.95& 0.6059&70.23\\
  \bottomrule
  \end{tabular}
  \caption{Ablation on the negative prompts.}
\label{tab:negative}
\end{subtable}

\begin{subtable}{\textwidth}
    \centering
     \begin{tabular}{c|ccccc|ccccc}
    \toprule
    \multirow{2}{*}{Pretrained Model}  &\multicolumn{5}{c|}{\textit{DrealSR}} & \multicolumn{5}{c}{\textit{RealSR}} \\
    & SSIM$\uparrow$ &LPIPS$\downarrow$& FID$\downarrow$ & MANIQA$\uparrow$ & MUSIQ$\uparrow$
    & SSIM$\uparrow$  &LPIPS$\downarrow$&FID$\downarrow$ & MANIQA$\uparrow$ & MUSIQ$\uparrow$ \\
    \midrule
    stable-diffusion-2-1&0.7378&0.3632 &155.09 & 0.4910&63.79&0.6912 &0.3329& 138.12& 0.5148&67.10\\
    stable-diffusion-1.5&0.7220&0.3864 &164.68 & 0.5713&67.84&0.6870 &0.3517& 141.95& 0.6059&70.23\\
  \bottomrule
  \end{tabular}
  \caption{Ablation on the base T2I diffusion models.}
    \label{tab:basemodel}
\end{subtable}
\end{table}

We verify different forms of constraints. As shown in \cref{tab:DFC}, both pixel-level and latent-level constraints contribute to generating more realistic images, resulting in improvements across multiple metrics. We claim that these two constraints aid in preventing the generation of confusing and incomprehensible content. 
As depicted in \cref{fig:sfa+dfc}~(b), the application of DFC significantly reduces image noise and alleviates the presence of aliasing artifacts. The output images exhibit textures with smooth edges, highlighting the effectiveness of DFC.

\paragraph{Effectiveness of negative prompts.}
We validate the impact of negative prompts, as described in \cref{sec:3.6}, by setting negative prompts as null text. In \cref{tab:negative}, it significantly improves the perceptual quality based on MAINIQA and MUSIQ. We also provide some visual results in \cref{fig:prompt+negative}~(b). 
By applying negative prompts, the images exhibit clearer textures, resulting in more realistic and higher quality.

\paragraph{Different base diffusion models.}
To explore the impact of the generative prior in the pre-trained T2I model, we analyze different versions of StableDiffusion\footnote{https://huggingface.co/stabilityai/stable-diffusion-2-1} in \cref{tab:basemodel}. 
Observations indicate a performance drop when employing SD-v2.1.
We speculate that this is due to a mismatch in the pre-training resolution, where SD-v2.1 is $768\times 768$ (different from our setting) and SD-v1.5 is $512\times 512$. However, this does not imply that the pre-training model has achieved its saturation point. \textbf{XPSR is orthogonal to these foundational models.} As the industry introduces superior base models, further improvements can be explored.

\section{Conclusion}

To address the challenge of accurately restoring semantic details in ISR tasks, we propose the XPSR framework.
We explore the role of different semantic priors in ISR and propose leveraging MLLMs to extract accurate cross-modal semantic priors as textual prompts.
To facilitate the integration of semantic priors, the SFA module is proposed.
To retain semantic-preserved information from degraded images, the DFC is applied.
Extensive experiments validate the strong performance of XPSR in generating images with clear details and semantic fidelity.
We hope this work can inspire exploring the cross-model priors in ISR.


%
%
\bibliographystyle{splncs04}
\bibliography{main}

\makeatletter
\renewcommand*{\@fnsymbol}[1]{\ifcase#1\or\dag\or\ddag\or*\or**\or***\or****\or*****\or******\or*******\or********\or*********\else\@ctrerr\fi}
\makeatother

\title{XPSR: Cross-modal Priors for Diffusion-based Image Super-Resolution} 
\subtitle{ ------ Supplementary Material ------}
\titlerunning{XPSR}

\author{Yunpeng Qu\inst{1,2}\thanks{Equal contribution.}\orcidlink{0009-0006-9700-6290}, Kun Yuan\inst{2}$^{\dag}$\thanks{Project leader.}\orcidlink{0000−0002−3681−2196}, Kai Zhao\inst{2}\orcidlink{0009-0003-5237-7512
}, \\
Qizhi Xie\inst{1,2}\orcidlink{0000-0001-6171-9789}, Jinhua Hao\inst{2}, Ming Sun\inst{2} \and Chao Zhou\inst{2}}

\authorrunning{Y Qu et al.}

\institute{Tsinghua University, China, Beijing \and Kuaishou Technology, China, Beijing\\
\email{\{qyp21,xqz20\}@mails.tsinghua.edu.cn}\\
\email{\{yuankun03,zhaokai05,haojinhua,sunming03,zhouchao\}@kuaishou.com}}

\maketitle

\section{Appendix}
\subsection{Implementation Details of XPSR}
\subsubsection{Image Encoder}

Stable Diffusion \cite{rombach2022high} utilizes a VAE Encoder to map the original $512\times 512$ image to a latent representation of $64\times 64 $ dimensions, enabling iterative denoising in the latent space.
To align with the scale of Stable Diffusion, the LR images are also upsampled by a factor of 4 to match the same resolution. 
Then, through a pyramid-like image encoder \cite{zhang2023adding}, the scale is gradually reduced to a $64\times 64$ feature space vector.
The image encoder consists of three layers, with each layer comprising two convolutional layers where the stride of the second convolution is $2\times 2$, resulting in a halving of the feature map size.

In our pixel-space constraint, we use linear layers to map the feature maps of $i$-th layer $\textbf{X}_i$ to RGB images $\hat{x}_i \in \mathbb{R}^{\frac{512}{2^i}\times\frac{512}{2^i} \times 3}$, which corresponds to the $i$-th scale of the HR image after downsampling. 
The $L_1$ loss is utilized to ensure that the extracted features closely resemble the semantic content of the HR image.

\subsubsection{ControlNet}
The ControlNet component, with a trainable copy of the Unet Encoder from Stable Diffusion, extracts multi-scale features through a pyramid structure. 
Within ControlNet, the features from LR images are further reduced from a dimension of $64\times 64$ to a latent representation of $8\times 8$.
The multi-scale conditional controls extracted by ControlNet are connected to the Unet through zero-convolution residual connections and \textit{conditional attention}.
Therefore, the conditional features at scale $j$ are also mapped to the channels of the latent representation through a linear transformation as $\hat{z}_j \in \mathbb{R}^{\frac{64}{2^j}\times\frac{64}{2^j} \times 4}$, corresponding to the $i$-th scale downsampled result of the HR latent.
The $L_1$ loss is also applied to enforce the constraint in the latent space.

\subsection{Additional Experimental Results}
\subsubsection{User Study}
\begin{table}[h]
  \centering
  \scriptsize
  \setlength{\tabcolsep}{5pt}
  \caption{Results of user study on real-world images.}
  \label{tab:userstudy}
  \centering
  \begin{tabular}{c|cccccc}
    \toprule
    Methods & BSRGAN & Real-ESRGAN  & StableSR & PASD & SeeSR & \textbf{XPSR(Ours)} \\
    \midrule
    Selection Rates & 0.7\% & 0.9\% &  5.3\% & 14.0\% & 14.8\% & \textbf{64.3\%}\\
  \bottomrule
  \end{tabular}
\end{table}

To thoroughly evaluate the performance of our XPSR in real-world scenarios, we conduct a user study on 50 LR real-world images randomly sampled from \textit{DrealSR} \cite{DBLP:conf/iccv/CaiZYC019} and \textit{RealSR} \cite{DBLP:conf/eccv/WeiXLZYZL20}. 
We compare our XPSR with five other ISR methods, including: BSRGAN~\cite{DBLP:conf/iccv/0008LGT21}, Real-ESRGAN~\cite{DBLP:conf/iccvw/WangXDS21}, StableSR~\cite{DBLP:journals/corr/abs-2305-07015}, PASD~\cite{yang2023pasd} and SeeSR \cite{DBLP:journals/corr/abs-2311-16518}.
For each image, the participants were simultaneously shown the LR image along with the restoration results from all ISR  methods, and they were then instructed to select the best ISR result for the LR image.
A total of 20 participants were invited to the user study and made a total of $20\times 50$ votes,  which are shown in \cref{tab:userstudy}.
Our method achieved \textbf{the highest selection rate of 64.3\%}, which is 4 times higher than the second-ranked method, showcasing the powerful application capabilities of XPSR in real-world scenarios.

\subsubsection{Comparisons on real-world images}

\begin{table}[t]
  \centering
  \tiny
  \caption{Quantitative comparison with SOTA methods on real-world dataset with no reference images.
  \textcolor{red}{Red} and \textcolor{blue}{blue} colors are the best and second-best performance.
  }
  \label{tab:headings}
  \centering
  \begin{tabular}{c|c|ccc|ccccc|c}
    \toprule
    \multirow{2}{*}{Dataset} & \multirow{2}{*}{Metrics} & \multicolumn{3}{c|}{GAN-based SR} & \multicolumn{6}{c}{Diffusion-based SR} \\
    & & BSRGAN & Real-ESRGAN & SwinIR & LDM & StableSR & DiffBIR & PASD & SeeSR & XPSR \\
    
    \midrule
    \multirow{3}{*}{\textit{RealLR200}}
    & MANIQA$\uparrow$   & 0.3671& 0.3633& 0.3741& 0.3049& 0.3688& 0.4288& 0.4295& \textcolor{blue}{0.4844}& \textcolor{red}{0.5589} \\
    & CLIPIQA$\uparrow$  & 0.5698& 0.5409& 0.5596&  0.5253&0.5935&0.6452 & 0.6325&\textcolor{blue}{0.6553} & \textcolor{red}{0.7524} \\
    & MUSIQ$\uparrow$    & 64.87& 62.96& 63.55& 55.19&  63.29 & 62.44 & 66.50& \textcolor{blue}{ 68.37} & \textcolor{red}{69.30} \\
  \bottomrule
  \end{tabular}
\end{table}

To evaluate the capabilities of our method in in-the-wild scenarios, we conduct tests on the \textit{RealLR200} dataset \cite{DBLP:journals/corr/abs-2311-16518}. The \textit{RealLR200} dataset consists of 200 real-world images, incorporating results collected from different studies \cite{DBLP:journals/corr/abs-2308-15070, DBLP:conf/iccvw/WangXDS21} as well as some images collected from the internet.
Due to the lack of available reference HR images for these real-world images, we only utilize three non-reference IQA metrics, including MANIQA \cite{DBLP:conf/cvpr/YangWSLGCWY22}, MUSIQ \cite{ke2021musiq}, and CLIPIQA \cite{wang2023exploring}. The quantitative results are shown in \cref{tab:headings}.

\begin{figure}[t]
  \centering
    \includegraphics[width=\linewidth]{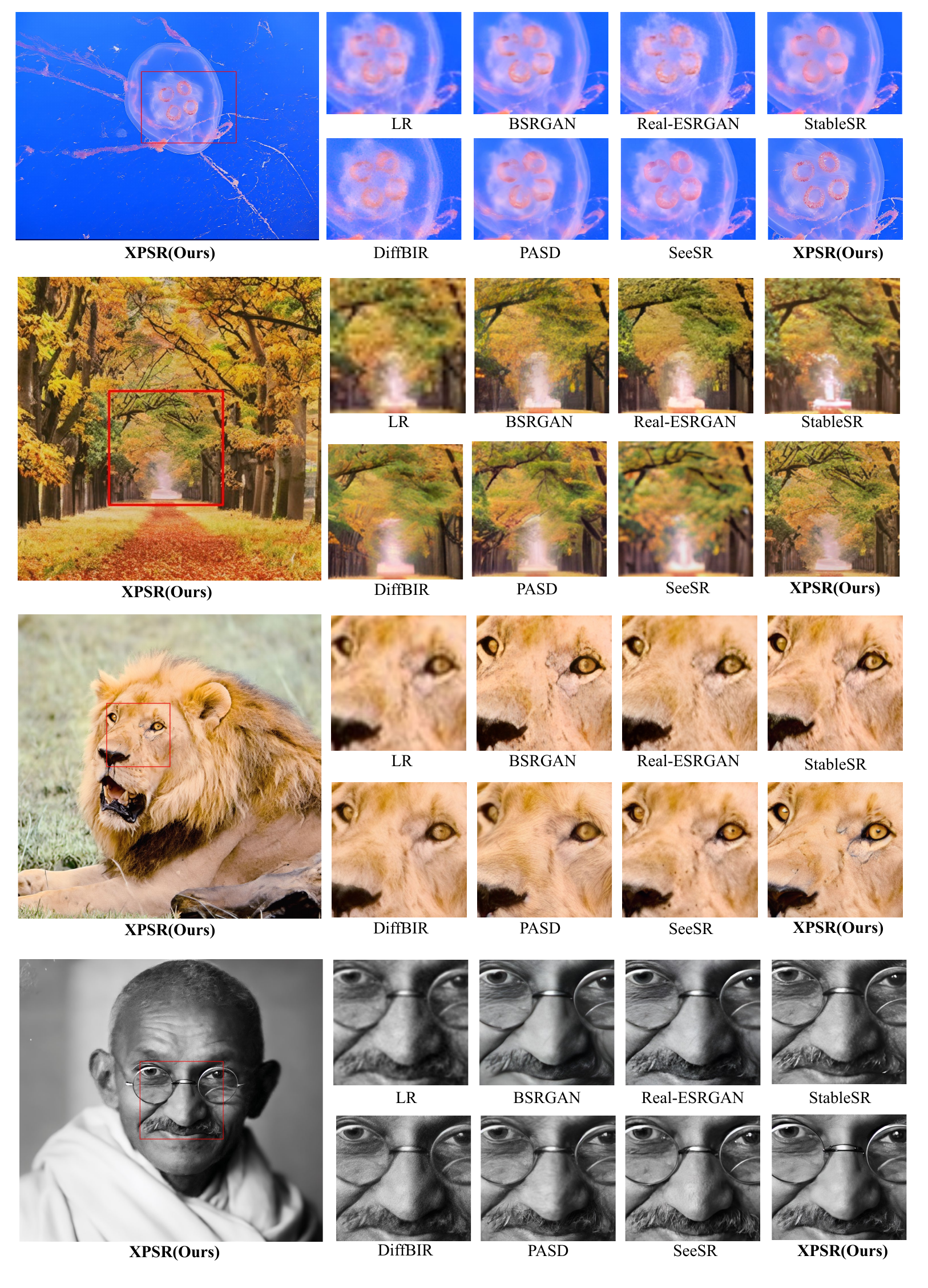}
  \caption{Qualitative comparisons with different SOTA methods on real-world images. \textbf{Zoom in for a better view}.}
  \label{fig:realimage}
\end{figure}

It can be observed that our XPSR performs \textbf{the best in all three metrics}, which is consistent with the results obtained on other datasets mentioned in the main text.
In addition, we have visualized some results in \cref{fig:realimage}, which indicate that XPSR is capable of recovering more realistic details compared to other methods, including lifelike facial features, intricate textures of the fur, and clearer tree leaves.
The aforementioned results clearly demonstrate the powerful image restoration capabilities of XPSR even in in-the-wild scenarios.

\begin{table}[t]
  \centering
  \tiny
  \caption{Ablation on the effect of MLLMs.}
  \label{tab:mllm}
   \centering
  \begin{tabular}{c|ccccc|ccccc}
    \toprule
     \multirow{2}{*}{Setting} &\multicolumn{5}{c|}{\textit{DrealSR}} & \multicolumn{5}{c}{\textit{RealSR}} \\
     & SSIM$\uparrow$ &LPIPS$\downarrow$& FID$\downarrow$ & MANIQA$\uparrow$ & MUSIQ$\uparrow$
    & SSIM$\uparrow$  &LPIPS$\downarrow$&FID$\downarrow$ & MANIQA$\uparrow$ & MUSIQ$\uparrow$ \\
    \midrule
    original  & 0.7220& 0.3864 & 164.68 & 0.5713 & 67.84 & 0.6870 & 0.3517 & 141.95 & 0.6059 & 70.23\\
    + LLaVA-13b & 0.7237 & 0.3870 & 165.70& 0.5760 & 67.94 &  0.6920 &  0.3508 & 141.94 &0.6099 &70.32 \\
    + Simple desc. & 0.7190  & 0.3859 & 168.10 & 0.5697  & 67.74  &0.6880 & 0.3537 &  148.93 & 0.5991 & 69.89 \\
    + Downscale 64 & 0.7212 & 0.3884 & 167.80  & 0.5706 &  67.44  &  0.6856 & 0.3588 & 144.54 & 0.5966 &  69.86 \\
    + Upscale 256 & 0.7204  &  0.3876 &  167.38 & 0.5733 &  67.89 &  0.6855 &0.3525 &142.45 &0.6079 &  70.36 \\
    + Gaussian blur & 0.7121 & 0.3926 & 175.77 & 0.5730  & 67.16  &  0.6812 & 0.3624 &  146.62& 0.5945 &  69.38 \\
    + Jpeg comp. & 0.7137& 0.3902 & 170.83  & 0.5690 & 67.42& 0.6822 &  0.3578 &  145.00 & 0.6040 &  69.70 \\
  \bottomrule
  \end{tabular}
\end{table}
\subsubsection{The impact of MLLMs on ISR performance.}
XPSR relies on MLLMs to obtain low-level and high-level semantic embeddings.
Therefore, it is necessary to explore how the limitations of MLLMs may impact the performance of the method.
We have explored based on the following three aspects. 
\textbf{(1)} \textit{Precision}. If an MLLM fails to understand LR, restoration results may be unrealistic or incorrect.
Hence, we utilize the larger and more accurate LLaVA-13b model to generate the prompt.
\textbf{(2)} \textit{Completeness}. As shown in Sec.3.3 of the paper, XPSR employs the detailed descriptions generated by the MLLMs as guidance.
For comparison, we build a simple prompt limited to 20 words, containing only object or distortion categories, to assess the impact of semantic completeness on image restoration.
\textbf{(3)} \textit{Robustness}. MLLM might struggle with diverse and complex image conditions, limiting its generalization ability. 
Therefore, we further degrade the LR images to obtain semantic cues under various complex degradation scenarios (\eg, resolution, JPEG compression, Gaussian blur).

As given in \cref{tab:mllm}, more precision and complete descriptions help to achieve better results, while the model suffers when LR owns severe degradations (\ie, JPEG compression and Gaussian blur). 
The resolution has little impact on the final result, which means that MLLM can still obtain accurate semantic descriptions.
\textbf{Notably, XPSR is orthogonal to MLLMs,} and better MLLMs in these three aspects can further advance XPSR.

\subsection{Limitations}

\begin{figure}[t]
  \centering
    \includegraphics[width=\linewidth]{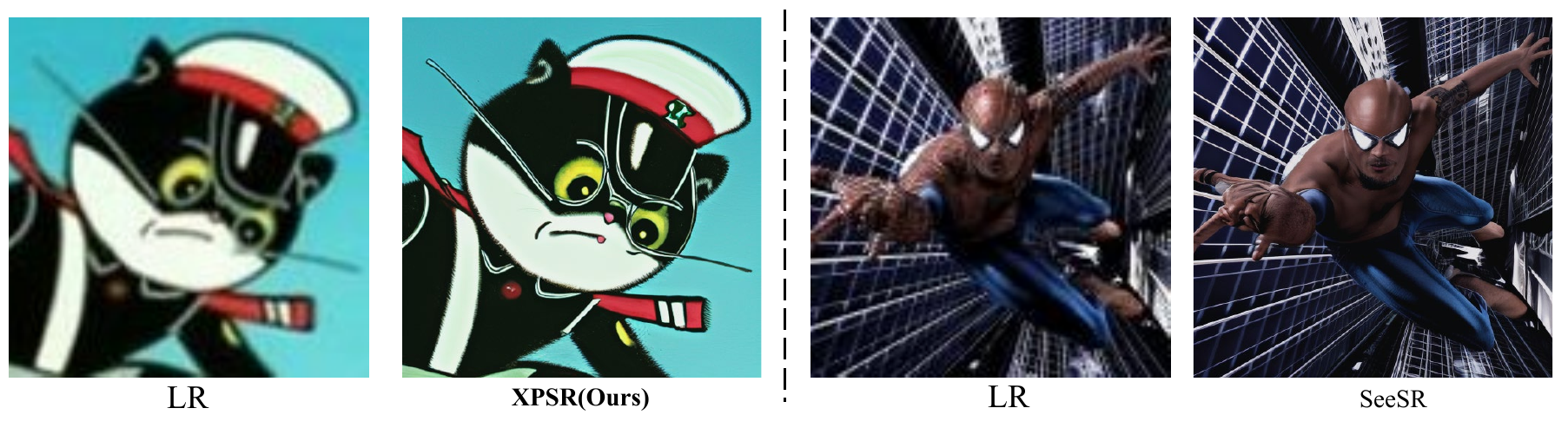}
  \caption{Limitations of diffusion-based methods. Due to their limited semantic understanding, the restored content may be unrelated to the original image.}
  \label{fig:limitation}
\end{figure}

In this section, we discuss the limitations of diffusion-based methods in the ISR task.
We have found that although diffusion models possess powerful generative capabilities and can produce realistic and detailed images, their understanding of specific scenes is limited. 
As a result, they can sometimes generate semantic-unrelated content.
In \cref{fig:limitation}, we present two examples. 

For the left case, although our XPSR model recognizes the main subject of the original image as a cat, it overlooks the cartoon-style nature of the image, resulting in the generated cartoon character exhibiting a realistic fur texture that is typically found only on real cats.
The same issue occurs in other diffusion-based models, as demonstrated in the case on the right. Although SeeSR recognizes the person's motion, it fails to realize that the original image depicts the famous character Spider-Man, restoring a clear but unrelated portrait.
Therefore, we consider that further enhancing the semantic understanding of different scenes is crucial for the successful application of diffusion models in ISR, which is also the motivation behind our XPSR. 
We strongly believe that this approach holds significant potential for exploration.

\subsection{More Visual Results}
\begin{figure}[t]
  \centering
    \includegraphics[width=\linewidth]{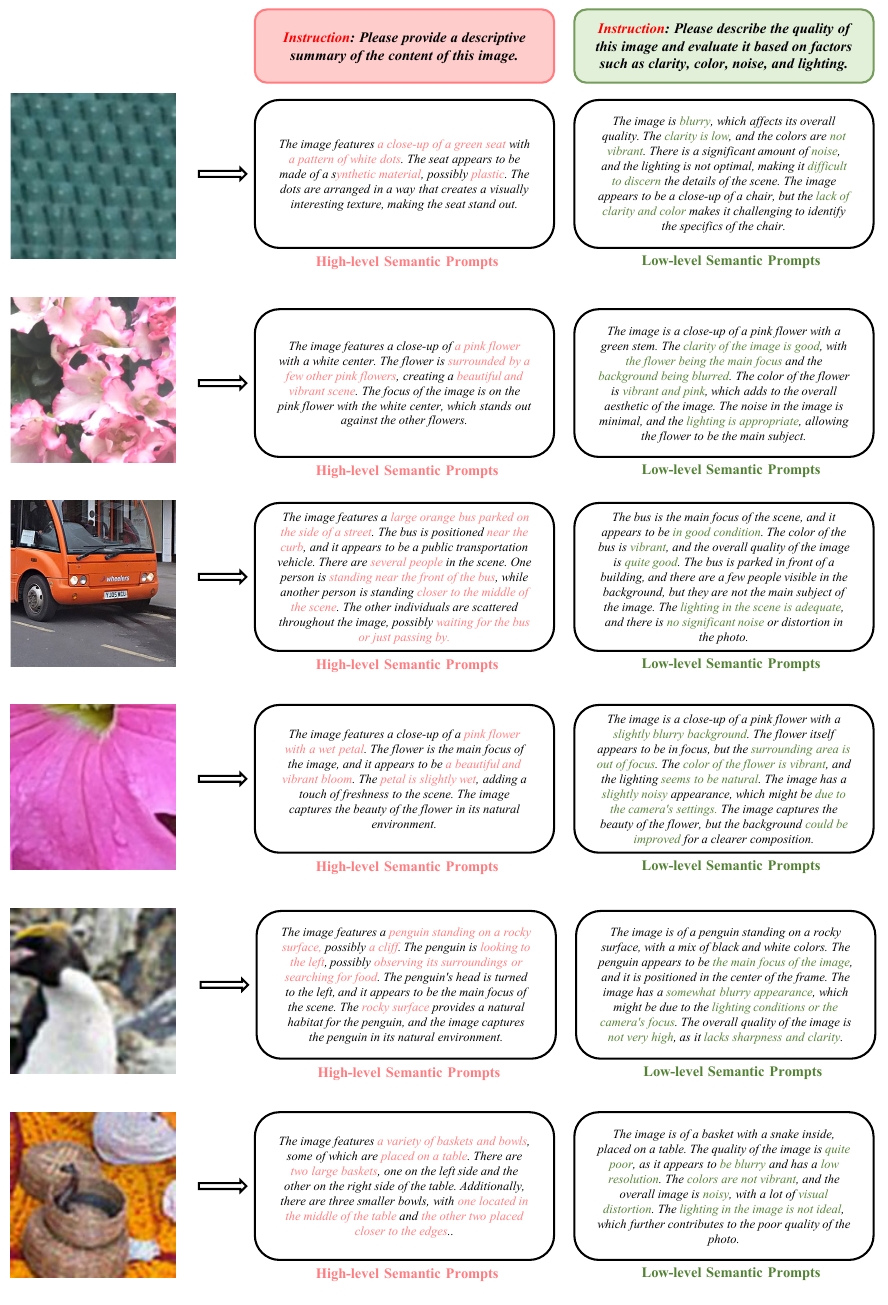}
  \caption{Additional cases show that LLaVA can generate high- and low-level semantic prompts consistent with human perception for both high- and low-quality images.}
  \label{fig:llava_add}
\end{figure}

\begin{figure}[t]
  \centering
    \includegraphics[width=\linewidth]{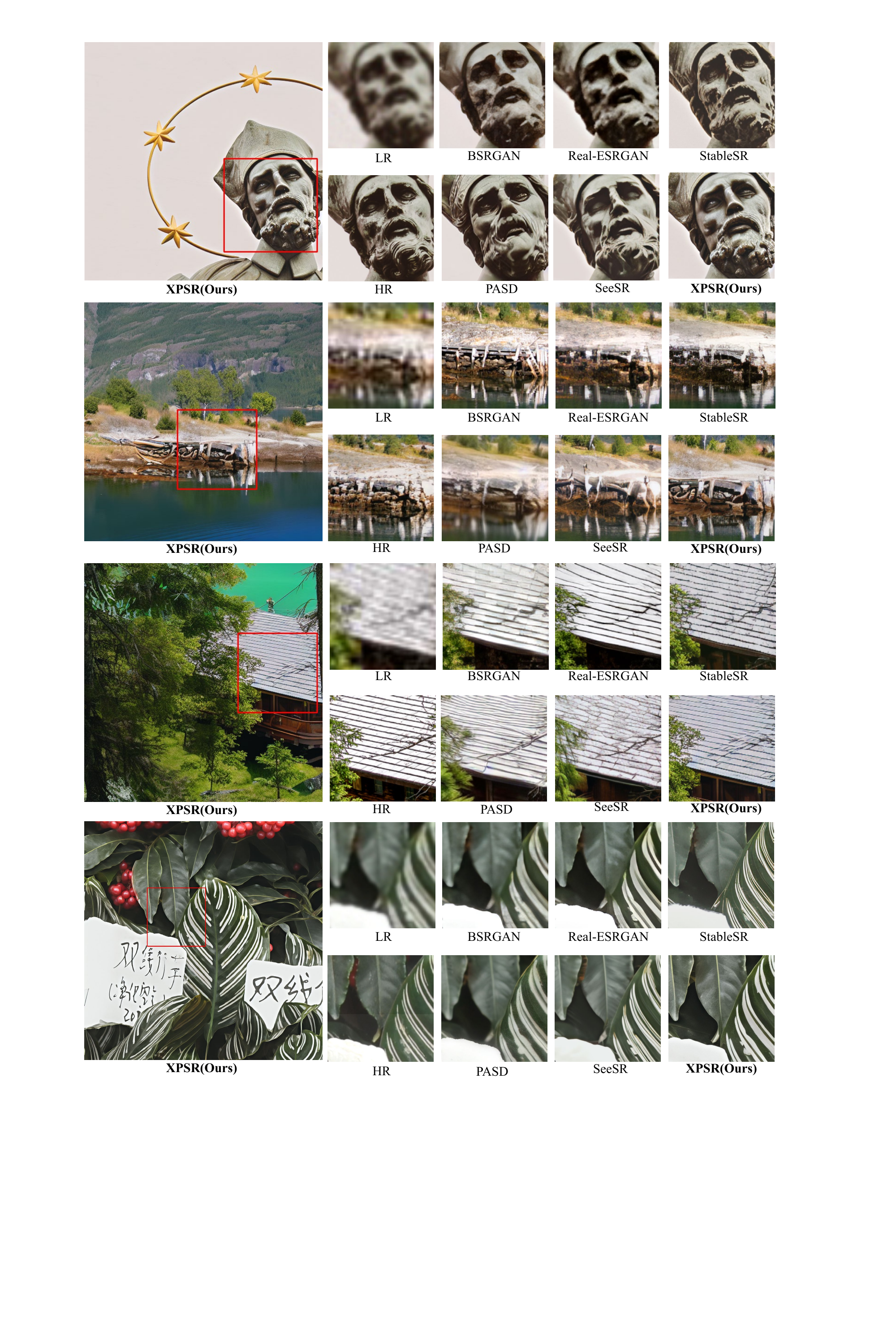}
  \caption{Additional qualitative comparisons with different SOTA methods (Part 1). \textbf{Zoom in for a better view}.}
  \label{fig:sota_add0}
\end{figure}

\begin{figure}[t]
  \centering
    \includegraphics[width=\linewidth]{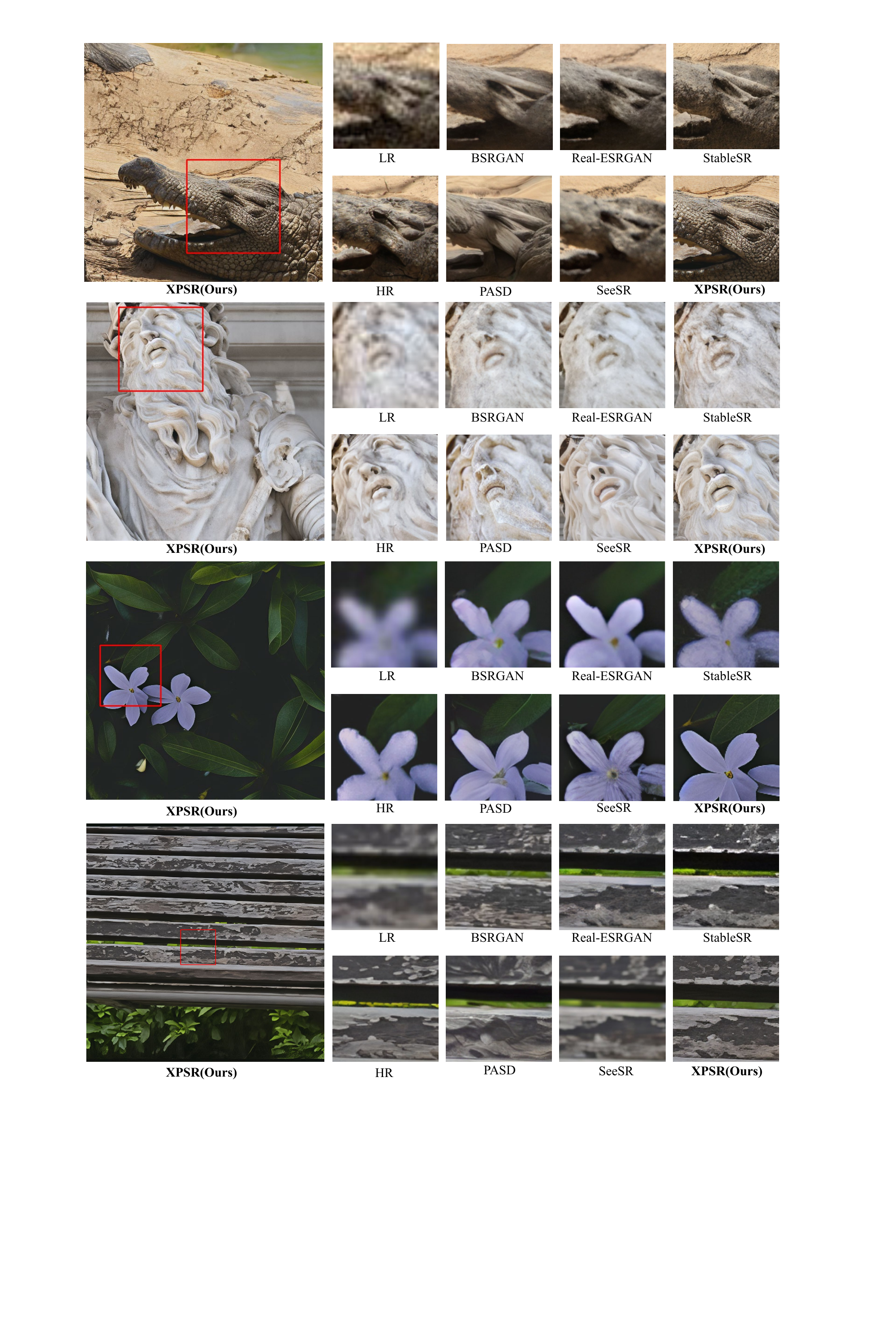}
  \caption{Additional qualitative comparisons with different SOTA methods (Part 2). \textbf{Zoom in for a better view}.}
  \label{fig:sota_add1}
\end{figure}

In this section, we provide additional visualization results. 
\cref{fig:llava_add} displays the high-level and low-level semantic prompts generated by LLaVA for different images, which align well with human perception. 
This illustrates the reliability of MLLM in incorporating cross-model semantic priors.
In \cref{fig:sota_add0} and \cref{fig:sota_add1}, we present additional comparative results with other methods, which further demonstrate the powerful capabilities of XPSR in generating high-fidelity and high-realistic images.
\clearpage

%
\end{document}